\documentclass[conference]{IEEEtran}
\IEEEoverridecommandlockouts
\usepackage{cite}
\usepackage{amsmath,amssymb,amsfonts}
\usepackage{algorithmic}
\usepackage{graphicx}
\usepackage{textcomp}
\usepackage{xcolor}
\usepackage{xspace}
\usepackage{booktabs}
\usepackage{multirow}
\usepackage{hyperref}
\def\BibTeX{{\rm B\kern-.05em{\sc i\kern-.025em b}\kern-.08em
  T\kern-.1667em\lower.7ex\hbox{E}\kern-.125emX}}
\begin{document}

\title{Adversarial Robustness of Neural-Statistical Features in Detection of Generative Transformers}

\author{\IEEEauthorblockN{Evan Crothers}
\IEEEauthorblockA{\textit{University of Ottawa}\\
Ottawa, Canada \\
ecrot027@uottawa.ca}
\and
\IEEEauthorblockN{Nathalie Japkowicz}
\IEEEauthorblockA{\textit{American University}\\
Washington DC, USA \\
japkowic@american.edu}
\and
\IEEEauthorblockN{Herna Viktor}
\IEEEauthorblockA{\textit{University of Ottawa}\\
Ottawa, Canada \\
hviktor@uottawa.ca}
\and
\IEEEauthorblockN{Paula Branco}
\IEEEauthorblockA{\textit{University of Ottawa}\\
Ottawa, Canada\\
pbranco@uottawa.ca}
}

\bstctlcite{IEEEexample:BSTcontrol}
\newcommand\bertbase{BERT$_{\small \textsc{BASE}}$\xspace}
\newcommand{\mauve}{{\fontfamily{bch}\selectfont{\textsc{Mauve}}}\xspace}

\makeatother

\maketitle

\begin{abstract}
The detection of computer-generated text is an area of rapidly increasing significance as nascent generative models allow for efficient creation of compelling human-like text, which may be abused for the purposes of spam, disinformation, phishing, or online influence campaigns. Past work has studied detection of current state-of-the-art models, but despite a developing threat landscape, there has been minimal analysis of the robustness of detection methods to adversarial attacks. To this end, we evaluate neural and non-neural approaches on their ability to detect computer-generated text, their robustness against text adversarial attacks, and the impact that successful adversarial attacks have on human judgement of text quality. We find that while statistical features underperform neural features, statistical features provide additional adversarial robustness that can be leveraged in ensemble detection models.  In the process, we find that previously effective complex phrasal features for detection of computer-generated text hold little predictive power against contemporary generative models, and identify promising statistical features to use instead.  Finally, we pioneer the usage of $\Delta$MAUVE as a proxy measure for human judgement of adversarial text quality.
\end{abstract}


\section{Introduction}

Generative text models capable of producing human-like text are a rapidly developing area of deep neural network development. The release of Generative Pretrained Transformer 2 (GPT-2) \cite{radford2019language} -- a model capable of high-quality unsupervised text generation -- was accompanied from the beginning by concerns that the model might be abused for malicious purposes. This resulted in a careful release schedule, where the largest GPT-2 model, a 1.5 billion parameter variant, was withheld following publication. The paper in which the subsequent GPT-3 model was introduced was not accompanied by an open-source release at all \cite{brockman_2020}. The model instead is being offered as an API, and has been licensed exclusively to Microsoft \cite{microsoft_2020}.

While limiting access to models of GPT-3 scale may reduce the scope of abuse in the near term, the widespread availability of the 1.5 billion parameter GPT-2 model (and other implementations of generative pre-trained Transformers with similar scale, such as Grover \cite{DBLP:journals/corr/abs-1905-12616}) have already opened a number of new attack possibilities. Fine-tuned variants already exist for simulating human comments, such as the GPT-2 Reddit project, which simulates submissions and comment threads at a very large scale \cite{disumbrationist_2019}. Should someone use such a model, and direct its generation through the selection of the correct prompt -- potentially with an auxiliary model to perform filtering of generated samples -- the result could allow a threat actor to effectively generate enormous volumes of text that mimics regular online discourse while promoting a specific agenda. Such a system is already feasible with widely-available open-source machine learning models today.

The likely proliferation of high-capacity generative text models is further increased by the promise shown by model distillation for large Transformer models \cite{sanh2019distilbert}, which allows large neural network models to be distilled into a smaller network with comparable performance. This reduces the hardware requirements and barrier to entry for those attempting to abuse such technologies, and has already been demonstrated successfully with GPT-2 \cite{wolf-etal-2020-transformers}. Furthermore, as companies provide access to APIs that give third-party developers access to massive-scale models, there is increased possibility that some portion of users may abuse the model. Finally, the Transformer architecture \cite{vaswani2017attention} that underpins the current crop of state-of-the-art generative text models is well known, and any well-funded groups with access to sufficient computational power has the ability to recreate a model of similar scale. As such, the threat that these models may be exploited for the purposes of fake news articles, fraudulent reviews, phishing campaigns, and similar is likely to increase over time.

With the understanding that technologies for generating human-like text at large scale are not only possible, but are already becoming increasingly accessible, research has increased in the area of developing techniques for detection of text that has been written by a computer. While methods have been developed to improve detection of generative text models, with specific focus specifically on large neural network Transformer models such as GPT-2, there has yet to be an in-depth assessment of the robustness of these models to adversarial attacks. Past research has shown that neural network models have difficult-to-mitigate vulnerabilities to adversarial attacks \cite{42503}. Such attacks allow an attacker to force an erroneous result from a neural network model through subtle perturbation of the input. The imagery domain in particular has been the subject of extensive research on adversarial attacks in assessment of specific target models, with particular focus on domains such as self-driving car systems \cite{8844593} and facial identification systems \cite{Yang2020}.

Given the potential scale of automated spam or influence campaigns when powered by generative text models, it is likely that human moderation alone will not be adequate for accurate detection of computer-generated text. Algorithmic approaches based on current machine learning models for detection of computer-written text are therefore the most practical solution to the problem, and are likely to be adopted by major technology and social media companies. Attackers are likely well aware of the types of detection models that such companies might employ, and may realistically use adversarial attacks to attempt to undermine detection methodologies. This motivates our assessment of the resilience of detection methods against concerted efforts to evade them.

With this environment in mind, this work evaluates the robustness of modern techniques for detection of computer-generated text by testing their performance in the presence of adversarial attacks in the text domain. We assess not only state-of-the-art neural networks, but prior work on statistical features as well.  We demonstrate that while neural network features outperform statistical features, incorporation of statistical features may improve adversarial robustness against specific adversarial attacks.  As part of this process, we investigate the relative predictive power of previously-effective statistical features, and provide guidance for future work.  We finally offer an assessment of vulnerability to adversarial attack that incorporates a proxy measure of human judgement, $\Delta$\mauve, providing a more concrete comparison of how perturbed an adversarial input must be to cause an erroneous classification.

The remainder of this work is structured as follows. Section II covers background knowledge and related work for understanding this research. Section III covers the experiment methodology. Section IV covers datasets and preprocessing procedures.  Section V covers experiment settings.  Section VI includes results of experiments. Section VII includes discussion of the results.  Finally, conclusions are presented in Section VIII.

\section{Related Work}

\subsection{Unsupervised Text Generation with Neural Networks}

Since the advent of the Transformer architecture \cite{vaswani2017attention}, generative text models using this architecture have become a significant area of research. The GPT-2 architecture \cite{radford2019language} was of particular importance in the area of unsupervised text generation, as it demonstrated the capacity for generating a large variety of text with a greater ability to pass for human text than any previous model. This model (and its successor, GPT-3) operate by taking as input a sequence of tokens, and then continue the sequence using a large pretrained language model to produce a probability distribution for the next token which can be repeatedly sampled to generate text. Variations of this architecture have produced new ways to tailor generated text, such as CTRL \cite{keskar2019ctrl}, which introduces the concept of control codes that affect the type of generated text beyond the initial prompt to the model.

Generative text models have well-understood potential for abuse in the context of spam, phishing, disinformation, and online influence campaigns. The original release of the 1.5 billion parameter version of GPT-2 was explicitly delayed based on concerns around the potential for abuse \cite{radford2019language}. The successor to this model, GPT-3, is an order of magnitude larger (175B parameters), with access only available via the official API \cite{brown2020language}. A sample of GPT-3 output is provided on the official GitHub repository of the model \cite{openai_2020}.


\subsection{Detection of Computer-Generated Text}

As the quality of models for generating text improves, the field of detection of computer-generated text has also grown. We group approaches for detection of computer-generated text as follows:

\begin{figure}
  \centering
  \includegraphics[width=0.5\textwidth]{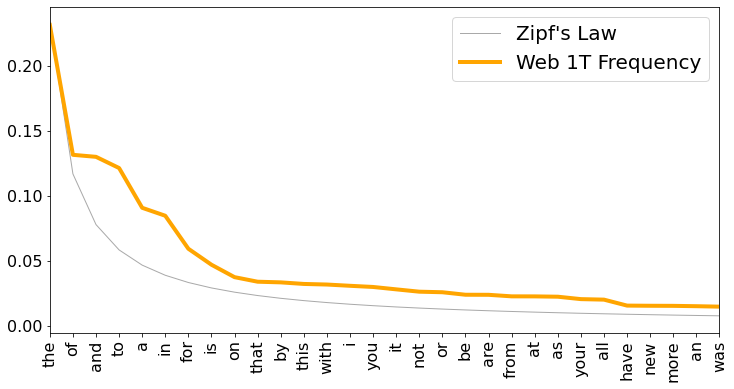}
  \caption{Normalized frequencies of 30 most common words in the Google Web Trillion Word Corpus\cite{web1t}\cite{norvig2009natural}, compared to theoretical Zipf's Law frequencies.}
  \label{fig:zipf}
\end{figure}

\subsubsection{Statistical Methods}

Computer-generated text does not always demonstrate the same statistical characteristics as human-generated text. For example, human-written text has been found to approximately conform with Zipf's Law -- the frequency of a word is inversely proportional to its rank in an ordering of words by frequency \cite{zipf1999psycho}\cite{zipf1949human}. The normalized frequency $f$ of a word of rank $k$ out of $N$ different words thereby follows the relationship:

\begin{align}
f\approx\frac{1/k}{\sum\limits_{n=1}^N (1/n)}
\label{eq:zipf}
\end{align}

In addition to the mathematical representation in equation \ref{eq:zipf}, a visual depiction of Zipf's Law compared to the frequency of words within the Google Web Trillion Word Corpus \cite{web1t} can be found in Figure \ref{fig:zipf}. This figure demonstrates that while there is typically some divergence between Zipf's Law and real-world word frequencies, there is nevertheless a Zipfian pattern in word frequency distribution.

In computer-generated text, such a trend in the relative frequency of words is not always observed to the same degree as in human-generated text \cite{8282270}. Past research also found that human-generated text tended to include more complex phrases, and that it tended to be more consistent according to sentence-level and paragraph-level consistency metrics \cite{8282270}. These findings were demonstrated to produce an effective method of computer-generated text detection when applied to 100 books computer-translated from Finnish to English. It was not studied, however, how much text is required for this method to be effective. Random variance at the level of a single comment or restaurant review may provide too short a sequence length for such an approach to be useful. Furthermore, as this approach was only applied to text produced by a particular version of Google Translate, these methods have been untested against current state-of-art text generation networks.  Recent work has used a feature-based approach to detection and characterization of GPT-2, GPT-3 and Grover datasets using a variety of text features, but intentionally avoids modern neural language models in the analysis, and does not consider adversarial robustness \cite{featuredetection}.

\subsubsection{Neural Networks}

One of the well-known neural approaches to detection of text generated by Transformer neural networks is known as ``Grover" \cite{DBLP:journals/corr/abs-1905-12616}. Grover is itself a generative text model, with an identical Transformer architecture to the original GPT-2, with the difference of using nucleus sampling rather than $top$-$k$ sampling for selecting the next word during generation. Grover was trained specifically trained on a corpus of internet news articles known as the RealNews dataset.  Grover's authors demonstrated the model is adept at detecting its own generated text.  In this process, a classification token \texttt{[CLS]} is appended to the input text and the final output state vector for this token is used as the input to a linear layer of neurons that is used to classify text samples: a common approach in sequence classification using Transformer models \cite{47751}.

Analysis of detection of computer generated text has determined that a bi-directional Transformer model (RoBERTa) substantially outperforms Grover models of equivalent parameter size for detection of GPT-2 text \cite{solaiman2019release}.  Discriminators from generative models such as Grover do, however, demonstrate improved performance when trained against the output of a smaller architecture (e.g., the GPT-2 355M parameter variant) and tested against a larger architecture (e.g., the GPT-2 1.5B parameter variant) \cite{brown2020language}\cite{fagni2021tweepfake}. Grover has also been found to underperform compared to other fine-tuned Transformers when classifying computer-generated text generated by other models than Grover itself \cite{uchendu2020authorship}.  
This should be unsurprising -- it would be unreasonable to expect that the discriminator of a GPT-2 model trained on a specific news dataset (i.e., Grover's detector) would outperform a discriminator not limited to that domain.
Preliminary research into detection of computer-generated text has already confirmed that adversarial attacks are effective against Grover \cite{wolff2020attacking}\cite{gagianorobustness}. In light of this, despite Grover's association with detection of computer-generated text due to targeting a perceived high-risk niche (i.e., ``fake news"), it is unreasonable and unfounded to expect Grover to perform well as a general-purpose detector for computer-generated text in any other number of myriad domains.  Neural approaches for broad detection of computer-generated text should then focus on the broader landscape of neural language models, including the Transformer models currently composing state-of-the-art.

\subsubsection{Human-Assisted Methods}

In the approach used by Giant Language Test Room (GLTR) \cite{DBLP:journals/corr/abs-1906-04043}, a neural and statistical approach is combined with coloured highlighting that assists a human analyst is determining whether a piece of text was generated by a machine or a person. This method uses other neural network language models, namely Bidirectional Encoder Representations from Transformers (BERT) \cite{47751} and the 117M parameter variant of GPT-2, to determine the probability of each word appearing in the sequence according to these models. The central assumption necessary for the success of this approach is that sampling methods used by generative models are biased towards more frequently occurring words -- a practice that improves the fluency of the resulting text output, but provides features that can be identified by the model.

There is clear overlap between this approach and automated approaches, but the presence of a human analyst is a factor worth considering as it is important for attackers in practice to evade platform moderators and cyber defence teams. Furthermore, reduced text quality due to an adversarial perturbation may mean the intended recipient of the text is unable to understand the original intended meaning, or may identify the text as untrustworthy.  To account for this, we will calculate \mauve scores -- a text quality measure that aligns with human judgements \cite{pillutla-etal:mauve:neurips2021} -- both before and after the adversarial attack to determine perceived degradation of text quality to a human observer.

Within this research, we address all three of these computer-generated text detection methods in our assessment of robustness against adversarial attacks in the text domain.

\subsection{Adversarial Attacks}
\label{ssec:adversarial_attacks}

Adversarial examples are inputs to a machine learning algorithm that are intentionally tailored to cause an incorrect result, typically by performing an easily overlooked perturbation to the input \cite{43405}. Building robustness against adversarial examples is an important element of assessing vulnerabilities in machine learning systems.

There are several types of attack modes that rely on adversarial examples, namely \textit{poisoning} and \textit{evasion} attacks \cite{Xu2020AdversarialAA}.  In a poisoning attack, adversarial examples are fed to the target model as training data to compromise its performance. In an evasion attack, the adversarial example is targets the model during inference to cause a targeted or untargeted erroneous result \cite{10.1007/978-3-319-66399-9_4}.

Many modern malware, phishing, and spam detection algorithms leverage machine learning \cite{10.5555/3130379.3130417}\cite{10.1145/3052973.3053009}\cite{5352759}. Applications of adversarial examples to malware detection might include poisoning attacks by intentionally sending misleading training data to a known honeypot server, or evasion attacks by altering the traffic produced by the malware such that it evades malware detection algorithms \cite{10.1007/978-3-319-66399-9_4}.

\begin{figure}
  \centering
  \includegraphics[width=0.5\textwidth]{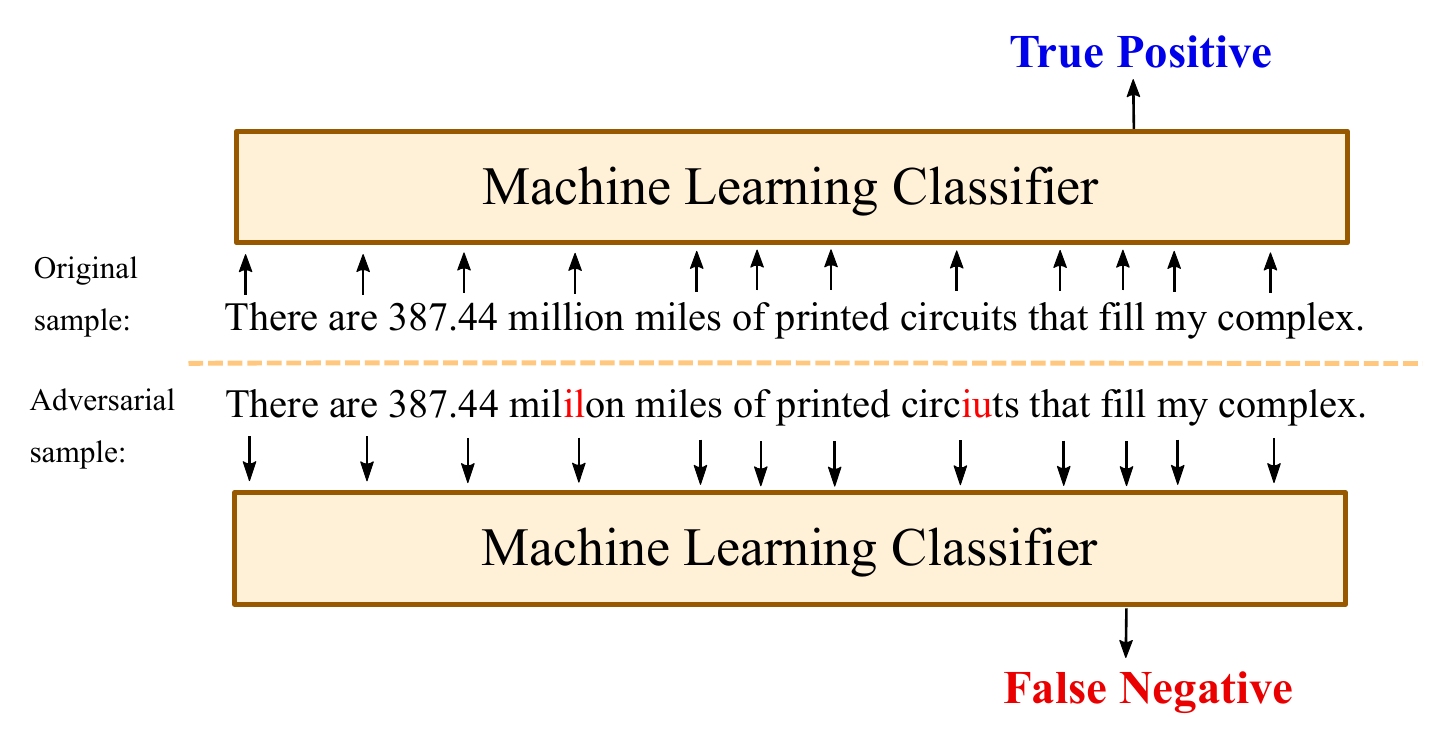}
  \caption{Example of how DeepWordBug can be used to trigger a Type II misclassification of computer-generated text via targeted character swaps \cite{gao2018black}. Targeted attacks for both Type I and Type II errors are considered in this work.}
  \label{fig:dwb}
\end{figure}

Due to the adversarial nature of cybersecurity, implementations of classifiers tuned to detect malicious content must be tested to evaluate their robustness against attempts to circumvent them. In many cases, even simple modifications to an input can be sufficient to fool a neural network classifier, such as affine transformations to images which are sufficient to cause a misclassification in many cases \cite{engstrom2019a}.

A major difference between adversarial examples in the image domain and adversarial examples in the text domain is that images provide continuous input data whereas the tokens in a text sequence are discrete. Furthermore, the output of a text generation model such as GPT-2 and GPT-3 is a probability distribution over the vocabulary of the model, from which the next word is sampled according to some sampling method (such as top-k or nucleus sampling). This sampling operation is non-differentiable and so it is not possible to propagate a gradient backwards across this step, typically preventing transferability of well-known attacks such as the fast gradient sign method (FGSM) to GPT-2 and GPT-3.

There has been significant recent work attempting to apply adversarial attacks to the text domain. A typical black-box approach is to replace words with synonyms in order to cause an erroneous classification, typically by leveraging some method of determining a synonym, or by leveraging another language model to find appropriate words given the context \cite{wang2019natural}\cite{alzantot-etal-2018-generating}. A more recent approach has leveraged the BERT model \cite{47751} to replace words with alternatives by masking out a particular word and filling by masking it and using the output of BERT to generate candidate replacements \cite{shi2019robustness}\cite{garg2020can}. This was found to improve the coherence of the generated sentence, and reduce the incidence of unnatural word replacements, in addition to improving the strength of the attack \cite{garg2020can}. A black-box attack known as DeepWordBug \cite{gao2018black} introduced targeted spelling errors to cause an erroneous classification, while maintaining text that a human can understand, and demonstrated this approach on spam classification algorithms.

Research has also been done to create adversarial examples in the text domain in the white-box setting, where the attacker has access to the model \cite{ebrahimi-etal-2018-hotflip}. By taking advantage of the gradients produced by the model when classifying a text sequence, an adversarial sample can be crafted by making a targeted character substitution, or ``flip". This white-box approach has the benefit of not requiring the generation of as many candidate perturbations, but only works on differentiable classification models and depends on the attacker having access to the model itself. As such, black-box attacks are often the more common threat model in application.

With an established variety of detection methodologies for computer-generated text, as well as a diverse range of methods for producing adversarial examples in the text domain, we now present a methodology for determining robustness of detection models against adversarial attacks. 

\section{Methodology}

In order to evaluate the classification performance and robustness of both statistical and neural classifiers for detection of computer-generated text, we follow two separate feature-extraction approaches. The first approach reflects statistical classification of computer generated-text in prior art \cite{8282270}, while the second represents a contemporary neural approach to the problem \cite{solaiman2019release}. Both are evaluated by testing their performance against computer-generated text created by generative pre-trained transformer (GPT) models of parameter counts 355M, 1.5B, and 175B respectively. During this analysis, we consider computer-generated text the positive class and human-written text the negative class. Following an assessment of each model's relative performance at classifying computer-generated text, the models are then evaluated for robustness in the presence of text adversarial attacks.

\subsection{Statistical Features}
\label{ssec:statmod}

The selection of statistical features is primarily based on past work that demonstrated 98.0\% accuracy in detecting machine generated text \cite{8282270}.  As this work was done prior to the advent of generative Transformer models, we are interested in assessing whether these features are still effective, and whether they are adversarially robust.  Notable among several categories of features (frequency features, complex phrasal features, and consistency features), previous work highlights the value of complex phrasal features \cite{8282270}. Complex phrasal features are based on the frequency of specific words and phrases within the analyzed text that occur more frequently in human text. To determine whether these features are still valuable, we harvest data from several online sources (listed in \S \ref{sec:preprocess}) to obtain a list of phrases that we then use calculate the complex phrasal features.

In selecting statistical features, we make three deviations from the work, which can be referenced for in-depth descriptions of the other features \cite{8282270}:

\begin{itemize}

\item We include two additional ``fluency" features: Gunning-Fog Index and Flesch Index. These features have been shown in recent work to be useful in detection of non-neural fake news when used in conjunction with LDA topic modeling \cite{10.1007/978-3-030-91434-9_29}, and provide a statistical measure of text readability and comprehensibility respectively.

\item When computing frequency features, we use a conventional mean-square error cost function when calculating information loss of a linear regression line that fits log-log lemma frequency versus rank. That is,

\begin{align}
\text{C}(y, \hat{y}) = \frac{1}{n} \sum_{i=0}^{n - 1} (y_i - \hat{y}_i)^2.
\label{eq:mse}
\end{align}

where $n$ is the number of distinct lemmas, $\hat{y}_i$ is the regression function evaluated for the lemma of rank $i$, $y_i$ is the true frequency of the lemma of rank $i$. 

\item We omit the complex phrasal feature of Yorkshire Dialect phrases. The feature had minimal predictive power in past research, and the dataset is not publicly available in computer readable form.

\end{itemize}

Besides these adjustments, other features are calculated using the same equations as in previous work \cite{8282270}, albeit with a complete reimplementation using modern NLP frameworks and with code made publicly available in the interest of reproducibility \cite{coderepo}.

Using the resulting statistical features, we train a support vector machine (SVM).  SVM models have been found to be the best performing model in previous computer-generated detection research using statistical features without neural networks \cite{8282270}\cite{featuredetection}.  Similar to past work, we train the SVM with a linear kernel.  In our case, the linear kernel provides two additional benefits: 1) output probabilities for integration with existing adversarial attack implementations, and 2) feature weights for interpretability of statistical feature importance.  A comparison to SVM models using RBF kernels confirmed comparable performance, with a top accuracy difference of less than 0.01 following exhaustive hyperparameter search across $C \in [1, 10, 100, 1000]$, $\gamma \in [0.001, 0.0001]$. 

\subsection{Neural Features}

To compare the robustness of statistical features to neural features, we leverage a selection of publicly available Transformer architectures pre-trained on several distinct datasets.  These models can be used to create a vector representation of an input sequence either through mean-pooling of output activations or taking the embedding of a special \texttt{[CLS]} token prepended to the input sequence, depending on the implementation \cite{reimers-2019-sentence-bert}.  The resulting feature vectors are used as input for an SVM classifier, trained in the same manner as discussed in \S \ref{ssec:statmod}.

The neural networks used as feature-extractors in this research are publicly available pre-trained networks provided in the Sentence Transformers repository for this purpose \cite{reimers-2019-sentence-bert}.  The feature-extraction approach enables direct comparison of features via a consistent inference algorithm, and allows for straightforward creation of ensembles via feature concatenation. This serves the overall goal of analyzing feature quality and adversarial robustness of statistical and neural features.

We are particularly interested in the widely-used RoBERTa architecture, as it forms the current state-of-the-art single model for detection of computer-generated text when fine-tuned specifically for detecting text from a particular generative model \cite{solaiman2019release}.  In addition to RoBERTa, we select four other pre-trained Transformer models from the Sentence Transformers project, selecting models with high task performance, trained on varying datasets, and of varying model sizes \cite{reimers-2019-sentence-bert}.  The Sentence Transformers documentation can be referenced for an explanation of how embeddings are calculated for each model.  Using features from pre-trained models reduces variation from separate fine-tuning processes, and enables reproducibility.  Every model evaluated in this work is publicly available with complete weights and can be found in the HuggingFace model repository \cite{wolf-etal-2020-transformers}.  Note that these feature extraction models were not pre-trained for detection of computer-generated text, and so do not represent an upper bound on overall classification performance.

\subsection{Evaluation Methodology}

We evaluate statistical and neural features on the ability of an SVM model trained as described in \S \ref{ssec:statmod} to correctly classify samples from 1) the WebText corpus used to train GPT-2 and GPT-3, and 2) computer-generated samples from trained GPT-2 and GPT-3 models. More effective features will exhibit higher accuracy and F1 scores on this classification task.  We leverage the official training and testing datasets provided by OpenAI for this purpose \cite{radford2019language}\cite{li_2020}.

To evaluate each model's robustness to text-based adversarial attacks, we subject each to DeepWordBug \cite{gao2018black} and TextFooler \cite{jin2019bert} adversarial attacks. We select these attacks as they are realistic black-box attacks that represent two disparate themes in text adversarial attacks: DeepWordBug causes small character edits and attempts to maximize misclassification while minimizing Levenshtein edit distance, while TextFooler replaces words based on a pre-trained bi-directional encoder representation from Transformers (BERT) model \cite{DBLP:journals/corr/abs-1810-04805}, replacing words with synonyms based on cosine similarity within the embedding space.

In attacking these models, we apply the assumption that the attacker has access to the output class confidence of the model, but not any internal weights or other model information.  As TextFooler and DeepWordBug attacks are quite expensive, especially given that some input sequences are quite long, we sample 200 random texts from GPT-2 355M texts and human-written WebText.  On this set, we perform targeted attacks for causing both Type I and Type II errors in computer-generated text detection.  To determine impact of adversarial attacks on text quality, we featurize \mauve against sentences from the original human-written samples in the WebText validation set.  Then, as there are too few successful DeepWordBug attacks to compute \mauve, we filter instances of successful TextFooler attacks against each model, and calculate \mauve scores of sentences from GPT-2 samples before and after the adversarial attacks are applied.  Finally, we calculate the resulting difference in \mauve score, $\Delta$\mauve, to determine the impact on human quality assessment.  As the number of sentences in comparatively small, we average $\Delta$\mauve over $k=10$ trials.  To our knowledge, this is the first time $\Delta$\mauve has been applied to assess impact of adversarial attacks on human-quality assessment.

\begin{table*}[!ht]
\centering
\caption{Performance of text feature embeddings for detection of computer-generated text}
\label{tab:result1}
\begin{tabular}{l|l|llllll|l|l}
\hline
\multirow{3}{*}{Features} & \multirow{3}{*}{Pre-Training Dataset} & \multicolumn{6}{c|}{SVM Accuracy / F1 Score}                                                                                                                                            & \multirow{3}{*}{C=} & \multirow{3}{*}{\begin{tabular}[c]{@{}l@{}}Feature\\ Size\end{tabular}} \\ \cline{3-8}
                          &                                       & \multicolumn{2}{c|}{GPT-2 355M}                                 & \multicolumn{2}{c|}{GPT-2 1.5B}                                 & \multicolumn{2}{c|}{GPT-3 175B}                    &                     &                                                                        \\ \cline{3-8}
                          &                                       & \multicolumn{1}{c}{Acc.} & \multicolumn{1}{c|}{F1}              & \multicolumn{1}{c}{Acc.} & \multicolumn{1}{c|}{F1}              & \multicolumn{1}{c}{Acc.} & \multicolumn{1}{c|}{F1} &                     &                                                                        \\ \hline
Statistical $(S)$              & N/A                                   & 0.7030                   & \multicolumn{1}{l|}{0.6935}          & 0.7120                   & \multicolumn{1}{l|}{0.7055}          & 0.5850                   & 0.5123                  & 100                 & 10                                                                    \\ 
RoBERTa $(R)$                  & 1B+ Weighted Web                      & 0.7700                   & \multicolumn{1}{l|}{0.7686}          & 0.7150                   & \multicolumn{1}{l|}{0.7053}          & 0.5660                   & 0.4694                  & 10                  & 1024                                                                   \\ 
Ensemble $(S+R)$                   & 1B+ Weighted Web                      & 0.8000                   & \multicolumn{1}{l|}{0.8008}          & 0.7450                   & \multicolumn{1}{l|}{0.7401}          & 0.6030                   & 0.5268                  & 10                  & 1034                                                                   \\ 
\midrule
MPNet \cite{10.1007/978-3-030-93046-2_42}                    & 1B+ Weighted Web                      & 0.7660                   & \multicolumn{1}{l|}{0.7715}          & 0.7290                   & \multicolumn{1}{l|}{0.7150}          & 0.5780                   & 0.4725                  & 10                  & 768                                                                    \\ 
MPNet \cite{10.1007/978-3-030-93046-2_42}                     & 215M QA Pairs                         & 0.7950                   & \multicolumn{1}{l|}{0.7968}          & 0.7450                   & \multicolumn{1}{l|}{0.7319}          & 0.5830                   & 0.4715                  & 1                   & 768                                                                    \\ 
\bertbase                 & MS MARCO                              & \textbf{0.8300}          & \multicolumn{1}{l|}{\textbf{0.8310}} & 0.7720                   & \multicolumn{1}{l|}{0.7645}          & 0.6200                   & 0.5343                  & 1                   & 768                                                                    \\ 
MiniLM \cite{NEURIPS2020_3f5ee243}                   & 215M QA Pairs                         & 0.7990                   & \multicolumn{1}{l|}{0.8012}          & \textbf{0.7860}          & \multicolumn{1}{l|}{\textbf{0.7780}} & \textbf{0.6520}          & \textbf{0.5807}         & 100                 & 384                                                                    \\ \hline
\end{tabular}
\end{table*}

\section{Datasets and Preprocessing}
\label{sec:preprocess}

To determine the suitability of the classifiers for detecting state-of-the-art neural generated text, we use the official dataset provided by OpenAI for assessing computer generated text from GPT-2 \cite{kim_2019}. This dataset contains GPT-2 generated samples created by GPT-2 networks of varying parameter counts. This dataset also includes samples from the original WebText corpus used to train the model, which can be used as negative examples for training the classifiers. Similarly, we also include a sample of GPT-3 output provided on the official OpenAI GitHub repository for GPT-3 \cite{openai_2020}. We train all models on a balanced training dataset of human webtext and GPT-2 355M output, and test on 3 separate test datasets balanced between human webtext and output from GPT-2 355M, GPT-2 1.5B, and GPT-3 respectively.  As an attacker may possess a large generative model that is not publicly available, it is useful to determine to what extent features derived from smaller architectures transfer to larger architectures.

In order to replicate the complex phrasal features used in past work in the field, we harvest three additional datasets from online repositories. This includes a dataset of cliché phrases \cite{hayden_1999}, a dataset of English idioms \cite{wikitionary}, and a dataset of Shakespearean archaisms \cite{cummings}. To collect this data, we scrape these web resources using Python scripts and assemble the data in text format. Where permissible, we have made the data available in easy-to-download format via GitHub repository. The remaining data can be provided upon request.

Preprocessing of the data is done using two separate preprocessing workflows designed for their respective models:

1) For the statistical model, we follow the approach used in past statistical detection of computer-generated text \cite{8282270}. We first tokenize the input using Stanza \cite{qi2020stanza}, then lemmatize the results. The number of tokens in the resulting text are tabulated as well. Lemmatization is also applied to the sets of complex phrasal features such that they can be matched against the lemmatized samples. Features are scaled by removing the mean and scaling to unit variance.

2) For Transformer models, data is first fed into a WordPiece tokenizer to convert it into tokens. Following this, the words are converted into identifiers based on their dictionary word IDs and provided to the neural network.

\section{Experimental Settings}

Experiments were executed on a virtual machine running Debian 10, with 32 vCPUs, 120GB RAM, and 4 NVIDIA T4 graphics processing units (GPUs). GPU acceleration was used to perform data preprocessing, model training, and deep learning inference more quickly whenever possible.

Throughout the experiments, hyperparameters are set to default values and random seeds are set to 0 to encourage reproducibility. All SVM models were trained with a C value based on exhaustive hyperparameter search across $C \in [1, 10, 100, 1000]$.  Models were trained using a linear kernel to allow class probability output and feature importance measurement of the statistical model.  Recall from \S \ref{ssec:statmod} that experiments using RBF kernels resulted in comparable accuracy results.

Text adversarial attacks were performed using the open-source TextAttack framework \cite{morris2020textattack}. In order to use this framework, we provide a harness to act as an adapter between the TextAttack libraries and the original models.

\section{Results}

We report the accuracies and F1 score of models trained on the statistical and neural features in Table \ref{tab:result1}. Results of adversary classification under the presence of TextFooler and DeepWordBug attacks can be found in Table \ref{tab:result2}.

\begin{figure}
  \centering
  \includegraphics[width=0.5\textwidth]{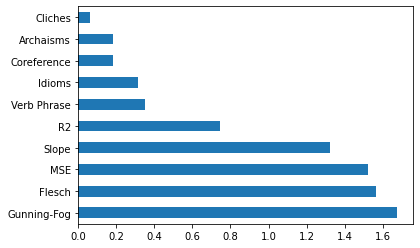}
  \caption{Feature weight comparison of SVM trained on statistical features}
  \label{fig:featimp}
\end{figure}

\begin{table*}[!ht]
\centering
\small
\caption{Demonstration of Textfooler adversarial attacks inducing Type I and Type II errors on web text}
\label{tab:tfsample}
\begin{tabular}{p{4.2cm}p{12.7cm}}
  \toprule
   \multicolumn{2}{c}{\textbf{Machine/Human Classification TextFooler Samples}} \\ \hline
   
   \textbf{Original (Label: Machine)} & Deejai Bhatt, who grew up in \textbf{\textit{Mississippi}} and moved to Memphis, says his time in America is an example of women being good in a bad environment.  \\ 
   \textbf{Adversary (Label: Human)} & Deejai Bhatt, who grew up in \textbf{\textit{Biloxi}} and moved to Memphis, says his time in America is an example of women being good in a bad environment.  \\ \hline
   
   \textbf{Original (Label: Human)} & The \textbf{\textit{protest}} events, scheduled to \textbf{\textit{take}} place just days before Trump takes office, will focus on the on-going effort to \textbf{\textit{repeal}} the Affordable Care \textbf{\textit{Act}}. \\ 
   \textbf{Adversary (Label: Machine)} & The \textbf{\textit{demonstrating}} events, scheduled to \textbf{\textit{adopt}} place just days before Trump takes office, will focus on the on-going effort to \textbf{\textit{abolishing}} the Affordable Care \textbf{\textit{Bill}}. \\
   \bottomrule
  \end{tabular}
  \label{tab:eg}
\end{table*}

\section{Discussion}

\subsection{Statistical feature importance}

Recall that as a linear kernel creates a separating plane within the same space as the input features, it is possible to use the coefficients of the SVM model as a measure of feature importance (\S \ref{ssec:statmod}).  A plotting of the weight of each statistical feature can be found in Figure \ref{fig:featimp}, and provides a summary of relative feature importance within the model.  The most heavily-weighted features in this analysis were the results of computation for Gunning-Fog Index and Flesch Index scores. These additional fluency features -- followed by Zipfian frequency features (slope, MSE, $R^2$) -- are of greatest importance in classifying computer-generated text produced by contemporary generative language models.

In contrast to past work that highlighted the efficacy of complex phrasal features \cite{8282270}, we find that complex phrasal features have low predictive power against Transformer-generated text compared to other features.  This is exhibited by the low weights attributed to these features (clichés, archaisms, and idioms) within the resulting statistical model (Figure \ref{fig:featimp}). In addition to improvements in underlying text generation models, this is likely due to shifts in the text domain considered. GPT-2 and GPT-3 are trained on web text, and produce text typically only as long as 10 paragraphs -- past work involved machine translation of book-length text. The presence of Shakespearian archaisms, writing clichés, and idioms is likely far more common in book text than in computer-generated blog posts and news articles. Of these complex phrasal features, idiom features retain the most predictive power in detection of current generative models.

\subsection{Classification performance}

In Table \ref{tab:result1}, we find that overall, features derived from pre-trained neural language models outperform the selected statistical features when attempting to classify larger language models using features from smaller models. Among neural language models, it appears that features from the comparatively compact MiniLM language model trained for a question-answering task are most amenable to maintaining stronger performance as the target model scales up, while features from the \bertbase model trained on MS MARCO web results perform well against the 355M parameter GPT-2 model, but do not transfer well to samples from larger architectures.

\subsection{Adversarial robustness}

Overall, we see from the results in Table \ref{tab:result2} that statistical features are considerably more robust to adversarial attack than Transformer-derived features.  An ensemble model trained on a vector of statistical and Transformer features provides a massive increase in adversarial robustness, while offering comparatively strong performance in non-attack settings.

We note that all models are more vulnerable to targeted attacks from TextFooler than DeepWordBug. Note that Textfooler performs a complete word exchange, whereas DeepWordBug makes small character-level changes in the form of dropping, adding, or swapping characters.  Excerpts from successful TextFooler attacks can be found in Table \ref{tab:tfsample}, an illustration of DeepWordBug can be found in Figure \ref{fig:dwb}.

Finally, within Table \ref{tab:result2} we see that all adversarial attacks reduce MAUVE scores.  When attacks succeed against statistical features, the resulting perturbed text demonstrates a greater decrease in \mauve score than successful attacks against RoBERTa features.  This indicates that attacks that succeed against the statistical model produce lower-quality text. A lower MAUVE score increases likelihood of detection upon human review, and further diminishes the ability of human targets to interpret the text. As an example, consider the TextFooler attacks shown in Table \ref{tab:tfsample}.  In the second passage, the words ``\textbf{\textit{take}} place" are perturbed to ``\textbf{\textit{adopt}} place", introducing a grammatical error. Further, the word ``Act" was perturbed to ``Bill", which while grammatically correct, semantically alters the meaning of the text. In all domains -- text, imagery, or otherwise -- a key goal of adversarial attacks is that they should influence machine interpretation, while maintaining human interpretation.

\begin{table}[]
\centering
\caption{Feature performance for computer-generated text detection in presence of adversarial attacks}
\label{tab:result2}
\begin{tabular}{lll|ll|l}
\hline
\multirow{2}{*}{Features} & \multirow{2}{*}{\begin{tabular}[c]{@{}l@{}}Attack\\ Type\end{tabular}} & \multicolumn{1}{l}{\multirow{2}{*}{\begin{tabular}[c]{@{}l@{}}Attack\\Succ. Rate\end{tabular}}} & \multicolumn{1}{l}{\multirow{2}{*}{\begin{tabular}[c]{@{}l@{}}Pre-Atk\\ Acc.\end{tabular}}} & \multicolumn{1}{l}{\multirow{2}{*}{\begin{tabular}[c]{@{}l@{}}Post-Atk\\ Acc.\end{tabular}}} & \multirow{2}{*}{$\Delta$\mauve} \\
                          &                                                                        & \multicolumn{1}{l}{}                                                                                 & \multicolumn{1}{l}{}                                                                        & \multicolumn{1}{l}{}                                                                         &                         \\ \hline
                          
    RoBERTa &  TF &                96.2\% &             0.780 &               0.030 &     -0.0115 \\
    RoBERTa & DWB &                61.5\% &             0.780 &               0.300 &     --  \\ \hline
Statistical &  TF &                \textbf{15.6\%} &             0.705 &               \textbf{0.595} &   -0.0947   \\
Statistical & DWB &                \textbf{3.5\%} &             0.705 &               0.680 &   --    \\ \hline
Ensemble &  TF &            27.8\%     &      \textbf{0.810}  &      0.585           &    -0.0364  \\
Ensemble & DWB &            8.0\%     &      \textbf{0.810}  &      \textbf{0.745}           &    --  \\
\end{tabular}

\end{table}

\section{Conclusion}

In analysis of feature robustness against adversarial attacks, we find that statistical features possess resistance against attacks that heavily impact neural language models.  This useful finding indicates that incorporation of statistical features may be a means of improving adversarial robustness of computer-generated text detection via ensemble models.

Statistical features previously useful in detection of computer-generated text -- specifically complex phrasal features -- are substantially less effective against contemporary models than in past research. Instead, additional features omitted from past work may be promising for this application, which include the Gunning-Fog and Flesch Indices.

\subsection{Future Work}

Improvements in quality of computer-generated text and the large variety of resulting threat models have created an adversarial cyber security environment.  As such, there is a broad need for defensive research into the robustness of detection methodologies, and methods for preventing widespread abuse of neural language models. Targeted research of detection within specific text domains (e.g., online comment sections) is also likely to be of value. 

\section{Acknowledgements}

This research was supported with Cloud TPUs from Google's TensorFlow Research Cloud (TRC).

\bibliographystyle{IEEEtran}
\bibliography{IEEEabrv,ref}

\end{document}